\documentclass[runningheads]{llncs}
\usepackage[T1]{fontenc}
\usepackage{graphicx}
\graphicspath{{./images/}}
\usepackage{bbding}
\usepackage[misc]{ifsym}
\usepackage[ruled,linesnumbered]{algorithm2e}
\usepackage[colorlinks=true, linkcolor=blue!60!cyan, citecolor=blue!60!cyan, urlcolor=blue!60!cyan]{hyperref}
\usepackage[table]{xcolor}
\usepackage{multirow}
\usepackage{mathrsfs}
\usepackage{booktabs}
\usepackage{threeparttable}
\usepackage{amsmath}
\usepackage{bbm}
\usepackage{bbding}
\usepackage{amssymb}

\begin{document}
\title{Holistic White-light Polyp Classification via Alignment-free Dense Distillation of Auxiliary Optical Chromoendoscopy}

%


\author{Qiang Hu$^{\dag}$\inst{1} 
	\and
	Qimei Wang$^{\dag}$\inst{1} 
	\and
	Jia Chen\inst{2} 
	\and
	Xuantao Ji\inst{2} 
	\and
	Mei Liu\inst{3} 
	\and
    Qiang Li \textsuperscript{\textrm{(\Letter)}} \inst{1} 
    \and
	Zhiwei Wang\textsuperscript{\textrm{(\Letter)}} \inst{1}} 

\titlerunning{Holistic White-light Polyp Classification}
\authorrunning{Q. Hu et al.}
%
\institute{Wuhan National Laboratory for Optoelectronics, Huazhong University of Science and Technology \and Changzhou United Imaging Surgical Co., Ltd \and Tongji Medical College of Huazhong University of Science and Technology\\
\email{\{huqiang77, qimei\_wang, liqiang8, zwwang\}@hust.edu.cn}}


\maketitle              
\def\thefootnote{$\dag$}\footnotetext{Equal~contribution; \textrm{\Letter}~corresponding author.}

\vspace{-0.1cm}
\begin{abstract}
White Light Imaging (WLI) and Narrow Band Imaging (NBI) are the two main colonoscopic modalities for polyp classification. While NBI, as optical chromoendoscopy, offers valuable vascular details, WLI remains the most common and often the only available modality in resource-limited settings. However, WLI-based methods typically underperform, limiting their clinical applicability. Existing approaches transfer knowledge from NBI to WLI through global feature alignment but often rely on cropped lesion regions, which are susceptible to detection errors and neglect contextual and subtle diagnostic cues.
To address this, this paper proposes a novel holistic classification framework that leverages full-image diagnosis without requiring polyp localization.
The key innovation lies in the Alignment-free Dense Distillation (ADD) module, which enables fine-grained cross-domain knowledge distillation regardless of misalignment between WLI and NBI images.
Without resorting to explicit image alignment, ADD learns pixel-wise cross-domain affinities to establish correspondences between feature maps, guiding the distillation along the most relevant pixel connections. To further enhance distillation reliability, ADD incorporates Class Activation Mapping (CAM) to filter cross-domain affinities, ensuring the distillation path connects only those semantically consistent regions with equal contributions to polyp diagnosis.
Extensive results on public and in-house datasets show that our method achieves state-of-the-art performance, relatively outperforming the other approaches by at least $2.5\%$ and $16.2\%$ in AUC, respectively.
Code is available at: \href{https://github.com/Huster-Hq/ADD}{https://github.com/Huster-Hq/ADD}.

\keywords{Colorectal polyps classification  \and Alignment-free distillation \and Cross-domain learning.}

\end{abstract}
\section{Introduction}
Colonoscopy is the gold standard for colorectal cancer (CRC) screening and diagnosis, where accurately distinguishing between \emph{adenomatous} and \emph{hyperplastic} polyps is essential~\cite{fonolla2020cnn}. Adenomatous polyps require removal due to their malignant potential, while hyperplastic polyps typically only necessitate periodic monitoring. However, variations in clinical experience can lead to inconsistencies in diagnosis. To enhance diagnostic objectivity, automated polyp classification algorithms for colonoscopy have been extensively studied~\cite{fonolla2019multi,tian2019one,ortega2021medical}.

Clinically, colonoscopy utilizes two primary imaging modes for polyp classification: white-light imaging (WLI), the standard modality, and narrow-band imaging (NBI), the optical chromoendoscopy modality. While NBI enhances vascular and morphological details by filtering specific wavelengths, WLI remains the predominant and only available option in resource-limited settings, such as capsule endoscopy. However, a significant performance gap exists between WLI- and NBI-based methods, with WLI generally exhibiting lower classification accuracy~\cite{usami2020colorectal,sierra2022deep}, limiting the broader adoption of polyp classification techniques. 

To address the accuracy gap and enhance the diagnostic precision of WLI, existing methods~\cite{wang2021colorectal,ma2022toward} have sought to leverage domain alignment strategies, knowledge from NBI to WLI to improve its discriminative capabilities. However, these approaches fundamentally rely on cropped lesion patches for polyp classification, thereby necessitating additional detection steps~\cite{hu2024sali,hu2025monobox} that may introduce systematic errors and hinder generalizability. Furthermore, by focusing solely on cropped lesion regions and employing global alignment techniques, these methods frequently overlook valuable contextual information (\emph{e.g.}, growth position and color contrast with surrounding tissue) and fail to adequately capture subtle diagnostic vascular information in the NBI domain.

In this paper, we propose a novel holistic classification framework designed to eliminate geometric priors while effectively integrating both lesion characteristics and contextual information through cross-domain dense knowledge distillation. However, WLI and NBI images are often misaligned due to temporal discrepancies, variations in viewing angles, and motion artifacts, rendering conventional fine-grained guidance from NBI to WLI unreliable.
To address this challenge, we introduce the innovative Alignment-free Dense Distillation (ADD) module, which rectifies dense knowledge distillation pathways from NBI to WLI along cross-domain pixel-wise affinities. These affinities identify the most relevant counterparts for each pixel between WLI and NBI feature maps, ensuring reliable knowledge transfer without requiring explicit spatial alignment overhead.
To further enhance distillation reliability, we incorporate Class Activation Mapping (CAM) to filter cross-domain affinities, constraining distillation pathways to link regions across domains that contribute equally to polyp diagnosis.

In summary, our major contributions are as follows: (1) We propose a novel holistic WLI polyp classification framework. To the best of our knowledge, this is the first exploration of leveraging NBI knowledge to assist holistic WLI polyp classifier without requiring polyp location. (2) We propose the Alignment-free Densen Distillation (ADD) module, which establishes dense distillation pathways between misaligned cross-domain features guided by learned affinities. Additionally, we capture the semantic relations to ensure distillation is restricted to semantically consistent regions. (3) Extensive experiments demonstrate that our method achieves the state-of-the-art performance in WLI image classification on both the public CPC-Paired and our in-house datasets, with relative  improvements of at least $2.5\%$ and $16.2\%$ in AUC, respectively.

\begin{figure}[t]
    \centering
    \includegraphics[width=0.95\linewidth]{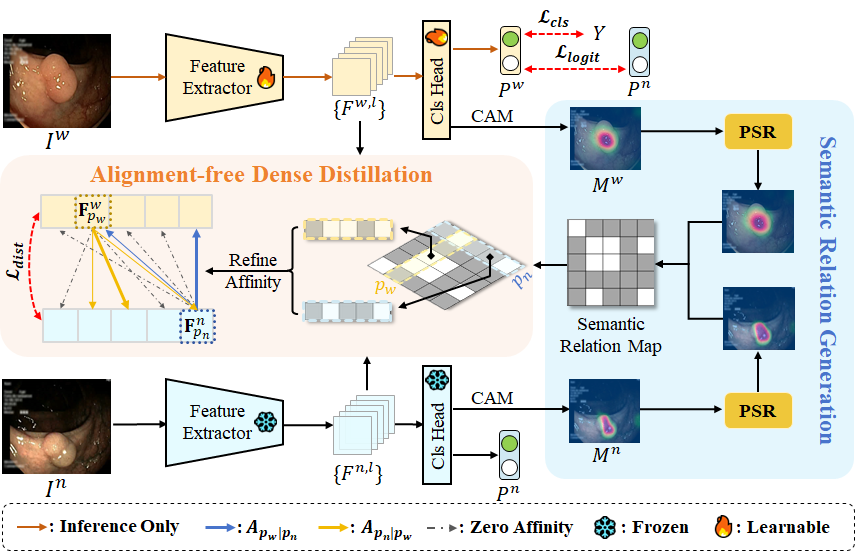}
    \caption{Overview of our framework, which focuses on enabling knowledge transfer from a frozen NBI classifier to a learnable WLI classifier. Its core module, Alignment-free Dense Distillation (ADD), adaptively establishes distillation pathways between misaligned cross-domain features, guided by bidirectional affinities. Furthermore, we introduce the Semantic Relation Generation (SRG) module, which generates a semantic relation map between cross-domain features based on the CAM maps optimized by Pairwise Similarity Refinement (PSR). This semantic relation map ensures that distillation is restricted to cross-domain feature vectors sharing the consistent semantic.}
    \label{fig:1}
\end{figure}

\section{Method}
As shown in Fig.~\ref{fig:1}, our framework consists of two classifiers: a WLI classifier and a well-trained NBI classifier.
During training, the NBI classifier remains frozen, while we introduce the Alignment-free Dense Distillation (ADD) module to facilitate efficient cross-domain knowledge transfer. ADD operates directly on the entire context-rich images, eliminating the need for explicit polyp location.

\subsection{Alignment-free Dense Distillation (ADD)}
Given paired but unaligned WLI-NBI images $(I^{w}, I^{n}) \in \mathbb{R}^{3 \times H \times W}$ and their shared ground truth (GT) category label $Y \in \{0, 1 \}$, where $0$ indicates hyperplastic and $1$ indicates adenomatous, we first pass $I^{w}$ and $I^{n}$ through the ResNet backbones of the learnable WLI classifier and the frozen NBI classifier to extract multi-scale feature maps, $\{ \mathbf{F}^{w,l},\mathbf{F}^{n,l} \} \in \mathbb{R}^{C_l \times H_l \times W_l}$, where $l$ denotes the resolution levels of intermediate feature maps. These feature maps are processed independently at each scale, with no cross-scale operations involved, and therefore, we omit the scale notation in the following sections.

In typical clinical acquisitions, $(I^{w}, I^{n})$ exhibit geometric misalignment due to endoscopic movement between modality switches. This spatial inconsistency causes cross-domain feature vectors at the same spatial location to potentially represent divergent anatomical regions, making their distillation suboptimal.
To address this issue, we propose the ADD module to establish dense distillation pathways guided by cross-domain affinities, which capture the relevance between cross-domain feature vectors. 

For clarity, we introduce notations $p$ and $k$ as spatial position index, and use the subscripts $w$ and $n$ to indicate which modality being indexed (\emph{e.g.}, $p_w$ for WLI and $p_n$ for NBI).
Given a WLI feature vector $\mathbf{F}^w_{p_w}$, its normalized affinity to an NBI feature vector $\mathbf{F}^n_{p_n}$ ($p_w$ can differ from $p_n$) can be formulated as follows:
\begin{equation}
    A_{p_n | p_w} = \frac{ \text{exp} ({S_{p_w,p_n}}) \cdot R_{p_w,p_n}}{ \sum_{k_n=0}^{H_lW_l-1} (\text{exp}(S_{p_w,k_n}) \cdot R_{p_w,k_n})},
    \label{eq:1}
\end{equation}
where $S_{p_w,p_n}$ denotes the cosine similarity between $\mathbf{F}^w_{p_w}$ and $\mathbf{F}^n_{p_n}$. Note that $R_{p_w,p_n}$ is a symmetric semantic relation between $\mathbf{F}^w_{p_w}$ and $\mathbf{F}^n_{p_n}$, which is used to filter out unreliable $S_{p_w,p_n}$. This will be explained in detail in Sec.~\ref{sec:2.3}.

To further handle domain discrepancies and better model the relationship between cross-domain feature vectors, we introduce a bidirectional affinity mechanism by simultaneously capturing the reverse affinity. That is, given $\mathbf{F}^n_{p_n}$, its affinity to $\mathbf{F}^w_{p_w}$ is defined as follows:
\begin{equation}
    A_{p_w | p_n} = \frac{\text{exp} ({S_{p_n, p_w})} \cdot R_{p_n,p_w}}{ \sum_{k_w=0}^{H_lW_l-1} (\text{exp}(S_{p_n, k_w}) \cdot R_{p_n,k_w})}.
    \label{eq:2}
\end{equation}

Both $A_{p_n | p_w}$ and $A_{p_w | p_n}$ can be seen as the probability of a distillation path existing between $\mathbf{F}^w_{p_w}$ and $\mathbf{F}^n_{p_n}$, but they are asymmetric since they are computed from different image modality perspectives. We apply a simple symmetrization operation by taking the union of both affinities, establishing symmetric dense distillation pathways between the WLI and NBI feature maps.
The dense distillation loss can then be expressed as follows:
\begin{equation}
\mathcal{L}_{dist} = \sum_{p_w=0}^{H_lW_l-1}\sum_{p_n=0}^{H_lW_l-1} (A_{p_n | p_w}+A_{p_w | p_n} - A_{p_n | p_w} \cdot A_{p_w | p_n}) \cdot \| \mathbf{F}_{p_w}^w - \mathbf{F}_{p_n}^n \|.
\label{eq:3}
\end{equation}
Here, every pixel between the unaligned WLI-NBI images can be connected, and knowledge is distilled from NBI to WLI according to the probability of path connectivity $(A_{p_n | p_w}+A_{p_w | p_n} - A_{p_n | p_w} \cdot A_{p_w | p_n})$.

\subsection{Semantic Relations Generation (SRG) for Refining Affinities}
\label{sec:2.3}
In addition to densifying and rectifying the distillation pathways, we ensure that distillation occurs only between cross-domain feature vectors that share the same semantic meaning. Inspired by works~\cite{chen2022class,ru2023token} using Class Activation Mapping (CAM)~\cite{zhou2016learning} to generate pseudo masks, we first generate CAM maps for the WLI and NBI images, denoted as $M^w$ and $M^n$, respectively, to roughly indicate polyp regions.
Next, we apply Pairwise Similarity Refinement (PSR) to iteratively refine these maps by aggregating pixel values within a window, guided by pairwise RGB-color similarities, for $T$ iterations. Taking $M^w$ as an example, at iteration $t$, the map is updated as follows:
\begin{equation}
    {M}_{p_w}^{w,t} =\sum_{k_w\in \mathcal{N} (p_w)}  \lambda_{k_w} \cdot M_{k_w}^{w, t-1}, \quad \lambda_{k_w} = \frac{1- \| I_{p_w}^{w} - I_{k_w}^{w} \|}{\sum_{k_w\in \mathcal{N} (p_w)} (1- \| I_{p_w}^{w} - I_{k_w}^{w} \|)},
\end{equation}
where $\mathcal{N} (\cdot)$ denotes the $3 \times 3$ neighboring pixel set.

Afterward, we resize the final refined CAM maps to the size of the feature maps. We then apply two thresholds, $\tau_1$ and $\tau_2$ ($0 < \tau_1 < \tau_2 < 1$), to binarize the maps, yielding $\mathbf{M}^{w}, \mathbf{M}^{n} \in \{0, \oslash, 1 \}^{H_l \times W_l}$, where $0$, $\oslash$, and $1$ represent the background, unsure, and polyp, respectively.
To determine whether the cross-domain pixels belong to the same class (excluding unsure pixels), the semantic relation between $\mathbf{F}^w_{p_w}$ and $\mathbf{F}^n_{p_n}$ is defined as follows:
\begin{equation}
    R_{p_w,p_n} = \begin{cases}
  1& \text{ if }~\mathbf{M}_{p_w}^w = \mathbf{M}_{p_n}^n ~\text{and}~\mathbf{M}_{p_w}^w,\mathbf{M}_{p_n}^n \ne \oslash\\
  0& \text{ otherwise }.
\end{cases}
\end{equation}
$R_{p_w,p_n}$ can be used to calculate $A_{p_n | p_w}$ and $A_{p_w | p_n}$ in Eq.(\ref{eq:1}) and Eq.(\ref{eq:2}), further pruning those unreliable WLI-NBI distillation paths connecting inconsistent CAM-derived semantics.

\subsection{Training details}
In addition to $\mathcal{L}_{dist}$, we compute additional loss terms: $\mathcal{L}_{logit}$ and $\mathcal{L}_{cls}$, to train the WLI classifier. Specifically, $\mathcal{L}_{logit}$ supervises the WLI classifier using the logit output of the NBI classifier as soft labels, while $\mathcal{L}_{cls}$ supervises the WLI classifier using the GT label $Y$. The overall training loss is formulated as:
\begin{equation}\small
\mathcal{L}_{total} = \mathcal{L}_{dist} + \mathcal{L}_{logit} + \mathcal{L}_{cls},
\label{eq:7}
\end{equation}
where $\mathcal{L}_{cls} = \mathrm{CE}(P^w, Y)$ and $\mathcal{L}_{logit}=\| P^w - P^n \|$.
Here, $\text{CE}(\cdot, \cdot)$ denotes the cross-entropy loss, and $P^w$ and $P^n$ represent the predicted logits of the WLI and NBI classifiers, respectively.
The implementation is by PyTorch~\cite{paszke2019pytorch} and the training is on a RTX 4090 GPU with 24GB memory. Weights of ResNet-50~\cite{he2016deep} pre-trained on ImageNet~\cite{deng2009imagenet}   are loaded as initialization. The input size is set to $448\times448$, and the batch size is set to $16$. The optimizer is set to Adam~\cite{kingma2014adam} with $1e-8$ weight decay. The number of epochs is set to $200$ and the initial learning rate is set to $1e-4$. We set $T$, $\tau_1$, and $\tau_2$ as $10$, $0.3$, and $0.7$, respectively, according to the parameter search experiment.

\section{Experiments}
\subsection{Datasets and Evaluation Metrics}
We evaluate our method on two datasets: the CPC-Paired dataset~\cite{wang2021colorectal} and our in-house dataset, both containing misaligned WLI-NBI polyp image pairs. The in-house dataset, collected from a local hospital, was annotated by two experienced colonoscopists. Both datasets consist of two polyp categories: hyperplastic and adenomatous. The CPC-Paired dataset includes $63$ hyperplastic and $102$ adenomatous pairs, while our in-house dataset contains $514$ hyperplastic and $598$ adenomatous pairs.

We perform 5-fold cross-validation on both datasets, ensuring patient-level separation between the train and test sets. To evaluate performance, we use six metrics: Accuracy (ACC), Precision (Pre), Sensitivity (Sen), Specificity (Spe), F1-score (F1), and Area Under the Curve (AUC). All evaluations are conducted on the WLI test set.

\subsection{Comparisons with State-of-the-art Methods}
\begin{table}[t]
\centering
\caption{Quantitative comparison on CPC-Paired dataset and our in-house dataset. The best performance is marked in bold and the second-best results are underlined.}
\fontsize{8}{9.3}\selectfont
\setlength{\tabcolsep}{0.6pt}
\label{tab:1}
\begin{tabular}{lcccccccccccc}
\toprule
\multirow{2}{*}{Method} & \multicolumn{6}{c}{CPC-Paired} & \multicolumn{6}{c}{In-house}\\
\cmidrule(lr){2-7}  \cmidrule(lr){8-13}
~&Acc~&Pre~&Rec~&Spe~&F1~&AUC~&Acc~&Pre~&Rec~&Spe~&F1~&AUC\\
\hline
SSL-CPCD~\cite{xu2024ssl} &0.709 &0.720 &0.833 &0.529 &0.772 &0.733 &0.577 &0.711 &0.485 &\textbf{0.711} &0.577 &0.642\\
SSL-WCE~\cite{guo2020semi} &0.750 &0.752 &0.863 &0.586 &0.804 &0.726 &0.523 &0.614 &0.530 &0.511 &0.569 &0.588\\
FFCNet~\cite{wang2022ffcnet} &0.860 &0.836 &0.951 &0.729 &0.890 &0.900 &0.622 &0.707 &0.621 &\underline{0.622} &0.661 &0.615\\
DLGNet~\cite{wang2023dlgnet} &0.872 &0.917 &0.863 &\underline{0.886} &0.889 &0.880 &0.649 &0.690 &0.742 &0.511 &0.715 &0.680\\
\hline
CPC-trans~\cite{ma2022toward} &0.895 &0.875 &\textbf{0.961} &0.800 &0.916 &0.882 &0.667 &0.644 &\textbf{0.985} &0.200 &\underline{0.779} &0.695\\
SAMD~\cite{shen2023auxiliary} &0.884 &0.902 &0.902 &0.857 &0.902&0.892&0.694&0.700&0.848&0.467&0.767&\underline{0.711}\\
PolypsAlign~\cite{wang2021colorectal} &\underline{0.907} &\underline{0.906} &\underline{0.941} &0.857 &\underline{0.923} &\underline{0.913} &\underline{0.712} &\underline{0.724} &0.833 &0.533 &0.775 &0.670\\
Ours &\textbf{0.930} &\textbf{0.959} &0.922 &\textbf{0.943} &\textbf{0.940} &\textbf{0.936} &\textbf{0.802} &\textbf{0.775} &\underline{0.939} &0.600 &\textbf{0.849} &\textbf{0.826}\\
\bottomrule
\end{tabular}
\end{table}
 
\begin{figure}[t]
    \centering
    \includegraphics[width=0.95\linewidth]{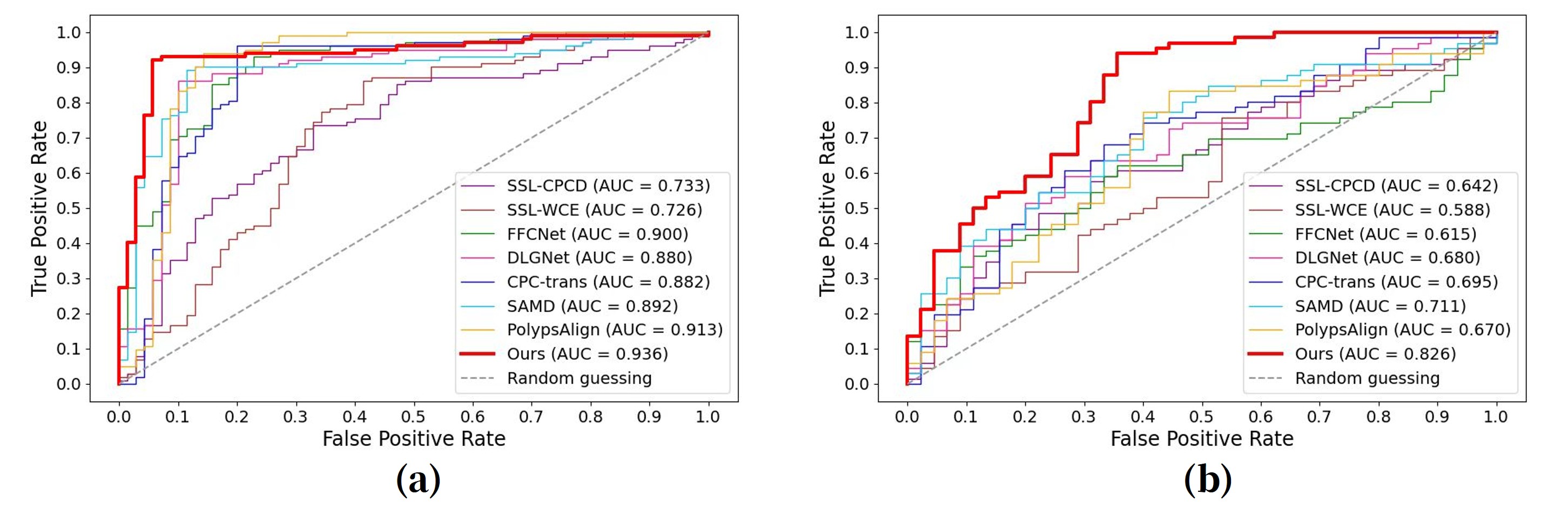}
    \caption{Comparison of ROC curves on (a) CPC-Paired and (b) our in-house dataset.}
    \label{fig:2}
\end{figure}
We compare our method with seven state-of-the-art (SOTA) methods: four cross-domain independent classification (CIC) methods~\cite{xu2024ssl,guo2020semi,wang2022ffcnet,wang2023dlgnet} and three cross-domain distillation classification (CDC) methods~\cite{wang2021colorectal,ma2022toward,shen2023auxiliary}. For CIC methods, we train models without any cross-domain interaction.
To ensure fairness, all methods use ResNet-50~\cite{he2016deep} as the backbone, except CPC-trans, which uses ViT-S~\cite{dosovitskiy2020image}. Notably, we implement CPC-trans~\cite{ma2022toward} and PolypsAlign~\cite{wang2021colorectal} to receive images cropped by GT polyp boxes during training and testing, following their original settings.

Table~\ref{tab:1} shows the quantitative comparison on the CPC-Paired and in-house datasets. CDC methods generally outperform CIC methods, confirming that knowledge transfer from the NBI domain to the WLI domain improves WLI classification. Our method achieves the best performance on most metrics for both datasets.
Surprisingly, our method, which receives holistic images, outperforms the two methods that receive GT polyp patches, which demonstrates the importance of global context for polyp classification.
Additionally, the ROC curves in Fig.~\ref{fig:2} show that our method achieves the highest True Positive Rates across most False Positive Rates. This improvement is due to our alignment-free dense distillation (ADD) module, which handles spatial misalignment and effectively reconstructs distillation pathways, unlike other CDC methods that rely on pixel-level alignment, missing complex spatial relationships across domains.

\subsection{Ablation Study}
\begin{table}[t] 
\centering
\caption{Ablation study of two key components: Alignment-free Dense Distillation (ADD) and Semantic Relation Generation~(SRG).}
\fontsize{8}{9.3}\selectfont
\setlength{\tabcolsep}{0.7pt}
\label{tab:2}
\begin{tabular}{cccccccccccccc} 
\toprule 
\multicolumn{2}{c}{Components} &\multicolumn{6}{c}{CPC-Paired} &\multicolumn{6}{c}{In-house}\\
\cmidrule(lr){1-2} \cmidrule(lr){3-8} \cmidrule(lr){9-14}
ADD &SRG &Acc &Pre &Rec &Spe &F1 &AUC &Acc &Pre &Rec &Spe &F1 &AUC\\
\hline 
\multicolumn{2}{c}{CIC variant} &0.721&0.776&0.745&0.686&0.760&0.801&0.568&0.650&0.591&0.533&0.619&0.603\\
\hline
\XSolidBrush &\XSolidBrush &0.826&0.891&0.804&0.857&0.845&0.857 &0.676&0.800&0.606&0.778&0.690&0.683\\
\Checkmark &\XSolidBrush &0.913 &0.922 &\textbf{0.931} &0.886 &0.926 &0.925 &0.766 &\textbf{0.885} &0.697 &\textbf{0.867} &0.780 &0.775\\
\Checkmark &\Checkmark &\textbf{0.930} &\textbf{0.959} &0.922 &\textbf{0.943} &\textbf{0.940} &\textbf{0.936} &\textbf{0.802} &0.775 &\textbf{0.939} &0.600 &\textbf{0.849} &\textbf{0.826}\\
\bottomrule
\end{tabular}
\end{table}

\begin{table}[t] 
\centering
\caption{Ablation study of two sub-strategies: Bidirectional Affinities and Pairwise Similarity Refinement, denoted as Bi-A and PSR, respectively.}
\fontsize{8}{9.6}\selectfont
\label{tab:3}
\begin{tabular}{lcccccccccccc}
\toprule
\multirow{2}{*}{Methods} &\multicolumn{6}{c}{CPC-Paired} &\multicolumn{6}{c}{In-house}\\
\cmidrule(lr){2-7}  \cmidrule(lr){8-13}
&Acc &Pre &Rec &Spe &F1 &AUC &Acc &Pre &Rec &Spe &F1 &AUC\\
\hline
Ours &\textbf{0.930} &\textbf{0.959} &0.922 &\textbf{0.943} &\textbf{0.940} &\textbf{0.936} &\textbf{0.802} &0.775 &\textbf{0.939} &0.600 &\textbf{0.849} &\textbf{0.826}\\
\hline
w/o Bi-A &0.913&0.931&0.922&0.900&0.926&0.918 &0.757&0.783&0.818&0.667&0.800&0.762\\
w/o PSR &0.919&0.907&\textbf{0.961}&0.857&0.933&0.928&0.775&\textbf{0.815}&0.803&\textbf{0.733}&0.809&0.786\\
\bottomrule
\end{tabular}
\end{table}

\subsubsection{Effectiveness of Key Components}
To evaluate the effectiveness of two key components, \emph{i.e.}, ADD and SRG, we train variants of our method by disabling ADD and/or SRG, and compare them with a variant trained in CIC manner.
Specifically, we disable ADD and SRG by replacing $\mathcal{L}_{dist}$ in Eq.~(\ref{eq:7}) with a consistency loss between paired WLI-NBI features, and train the CIC variant similarly to the methods in Table~\ref{tab:1}. Note that SRG is designed to assist ADD, so it cannot operate independently.
The comparison results in Table~\ref{tab:2} show that even without optimal feature-level distillation, aligning output logits alone can effectively improve classification accuracy for WLI images (comparing the first two rows).
Further analysis reveals that the ADD module significantly boosts performance, with AUC increasing by $0.068$ and $0.092$ on the CPC-Paired and in-house datasets, respectively (comparing the first three rows). 
This improvement stems from ADD's ability to establish adaptive distillation paths, effectively addressing spatial misalignment.
Finally, the SRG module further enhances performance by restricting distillation paths to features with semantic consistency, optimizing knowledge transfer.


\subsubsection{Effectiveness of Sub-strategies}
To assess the impact of sub-strategies, \emph{i.e.}, bidirectional affinities (Bi-A) and pairwise similarity refinement (PSR), we train variants with these components disabled. Specifically, we disable Bi-A by keeping only $A_{{p_n}|{p_w}}$ in Eq.~(\ref{eq:3}). As shown in Table~\ref{tab:3}, removing Bi-A or PSR results in varying performance degradation. Undirectional affinities fail to capture complex cross-modal relationships, while the absence of PSR introduces noise into the semantic relations from coarse CAM maps.

\subsection{Visualization of CAM}
\begin{figure}[t]\small
    \centering
    \includegraphics[width=0.9\linewidth]{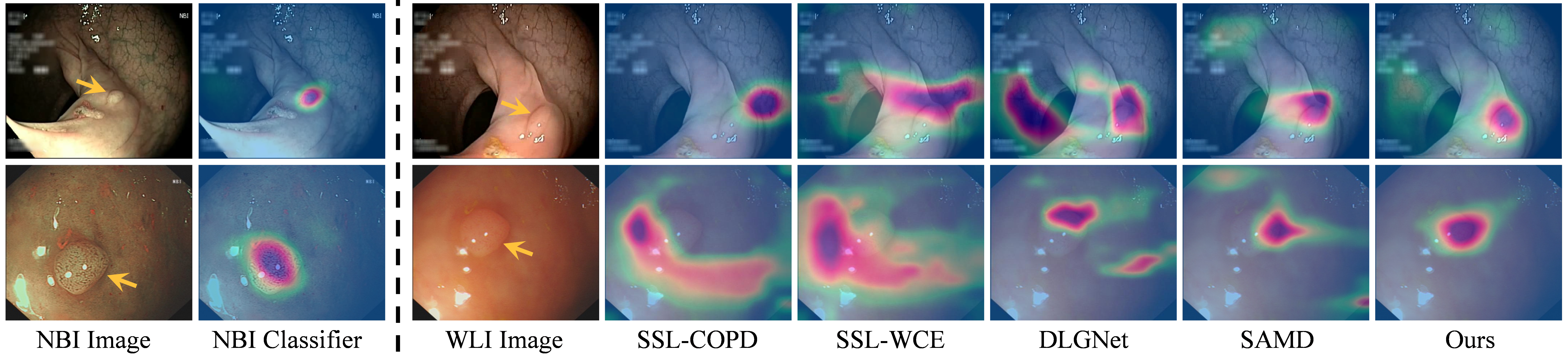}
    \caption{Visualization of CAM maps of different methods, which reflect the `attention region' while model classifying. Yellow arrows indicate polyps.}
    \label{fig:3}
\end{figure}
We visualize the CAM maps of different classifiers that receive holistic images in Fig.~\ref{fig:3}. CDC methods show more precise focus on polyp regions than CIC methods, confirming the importance of NBI knowledge distillation for WLI classification. Our method, in particular, produces attention regions that are most aligned with the NBI classifier, demonstrating the effectiveness of our cross-domain distillation and validating the use of CAM to guide relation map construction.

\section{Conclusion}
In this work, we propose a novel framework for WLI polyp classification, introducing the Alignment-Free Dense Distillation (ADD) module and Semantic Relation Generation (SRG) module. The ADD module enables effective knowledge distillation from the NBI domain to the WLI domain, while the SRG module ensures distillation occurs only between semantically consistent regions, optimizing the knowledge transfer. By addressing spatial misalignment and ensuring semantic alignment, our method significantly enhances WLI classification accuracy.
Experimental results on the CPC-Paired and in-house datasets show that our method outperforms state-of-the-art polyp classification methods, achieving superior performance. Visualization results further confirm that our method produces the most accurate attention regions, validating the alignment of cross-domain features and demonstrating the effectiveness of CAM-guided distillation.

\begin{credits}
\subsubsection{\ackname}
This work was supported in part by National Natural Science Foundation of China (Grant No. 62202189), Key R\&D Program of Hubei Province of China (No. 2023BCB003), and Changzhou United Imaging Surgical Co., Ltd.
Thanks to Professor Mei Liu's team for providing the in-house dataset used in this work at the Department of Gastroenterology, Tongji Medical College, Huazhong University of Science and Technology.


\subsubsection{\discintname}
The authors have no competing interests to declare that are relevant to the content of this article.
\end{credits}
\bibliographystyle{unsrt}  
\bibliography{reference}

\end{document}